%% file: example_paper.tex
\documentclass{article}

\usepackage{microtype}
\usepackage{graphicx}
\usepackage{subcaption}
\usepackage{booktabs} 
\usepackage{tcolorbox}
\tcbuselibrary{listings,breakable}

\usepackage{listings}
\usepackage{xcolor}
\usepackage{tabularx}
\usepackage{xspace}

\usepackage{hyperref}

\usepackage[preprint]{icml2026}

\usepackage{amsmath}
\usepackage{amssymb}
\usepackage{mathtools}
\usepackage{amsthm}

\usepackage[capitalize,noabbrev]{cleveref}

\theoremstyle{plain}

\theoremstyle{definition}

\theoremstyle{remark}

\def\method{World-Gymnast\xspace}

\usepackage[textsize=tiny]{todonotes}

\newcommand{\algcomment}[1]{\hfill $\triangleright$~#1}
\renewcommand{\algorithmicrequire}{\textbf{Require:}}
\renewcommand{\algorithmicensure}{\textbf{Ensure:}}

\icmltitlerunning{World-Gymnast: Training Robots with Reinforcement Learning in a World Model}

\begin{document}

\twocolumn[
  \icmltitle{World-Gymnast: Training Robots with \\ Reinforcement Learning in a World Model}



  \icmlsetsymbol{equal}{*}

  \begin{icmlauthorlist}
    \icmlauthor{Ansh Kumar Sharma}{nyu}
    \icmlauthor{Yixiang Sun}{nyu}
    \icmlauthor{Ninghao Lu}{nyush}
    \icmlauthor{Yunzhe Zhang}{nyu}
    \icmlauthor{Jiarao Liu}{ucb}
    \icmlauthor{Sherry Yang}{nyu}

    \vspace{2pt}
\vspace{4pt}
{\normalsize\centerline{
\href{https://world-gymnast.github.io}{world-gymnast.github.io}
}}
\vspace{-2pt}
  \end{icmlauthorlist}

  \icmlaffiliation{nyu}{New York University}
  \icmlaffiliation{nyush}{NYU Shanghai}
  \icmlaffiliation{ucb}{University of California, Berkeley}

  \icmlcorrespondingauthor{Ansh Kumar Sharma}{as20482@nyu.edu}
  \icmlcorrespondingauthor{Sherry Yang}{sherryyang@nyu.edu}

  \icmlkeywords{World Models, Model-Based Reinforcement Learning, Vision-Language-Action Models, Robot Learning}

  \vskip 0.3in
]



\printAffiliationsAndNotice{}  


\begin{abstract}  
  Robot learning from interacting with the physical world is fundamentally bottlenecked by the cost of physical interaction. The two alternatives, supervised finetuning (SFT) from expert demonstrations and reinforcement learning (RL) in a software-based simulator, are limited by the amount of expert data available and the sim-to-real gap for manipulation. With the recent emergence of world models learned from real-world video-action data, we ask the question of whether training a policy in a world model can be more effective than supervised learning or software simulation in achieving better real-robot performance. We propose \method, which performs RL finetuning of a vision-language-action (VLA) policy by rolling out the policy in an action-conditioned video world model and rewarding the rollouts with a vision-language model (VLM). On the Bridge robot setup, \method outperforms SFT by as much as 18x and outperforms software simulator by as much as 2x. More importantly, \method demonstrates intriguing capabilities of RL with a world model, including training on diverse language instructions and novel scenes from the world model, test-time training in a novel scene, and online iterative world model and policy improvement. Our results suggest learning a world model and training robot policies in the cloud could be the key to bridging the gap between robots that work in demonstrations and robots that can work in anyone's household.
\end{abstract}

\setlength{\textfloatsep}{0.5ex}


\input{introduction}
\input{preliminaries}
\input{method}
\input{experiment}
\input{related}
\input{conclusion}




\section*{Impact Statement}

This work introduces \method, a framework for training robot policies within a generative world model. Our approach has the potential to democratize robotic research by reducing the dependency on expensive physical hardware and manual simulator engineering, thereby lowering the barrier to entry for developing capable generalist robots. While this method significantly mitigates the physical risks and costs associated with real-world training, we acknowledge that reliance on generative video models introduces the risk of policies exploiting hallucinated physics. Consequently, despite the improved sim-to-real transfer demonstrated in our results, policies trained in such environments should undergo rigorous safety verification before deployment in safety-critical real-world settings.

\section{Acknowledgement}
We would like to thank the AutoEval~\citep{zhou2025autoeval} team, especially Zhiyuan (Paul) Zhou, for their assistance on conducting the AutoEval experiments. We would like to thank Julian Quevedo for his guidance on using the WorldGym~\citep{quevedo2025worldgymworldmodelenvironment} codebase.


\bibliography{example_paper}
\bibliographystyle{icml2026}

\newpage
\appendix
\onecolumn
\section{Additional Details of The Models and Training}
\subsection{Fast Inference on World Model}
A major bottleneck in the RL training pipeline is the world model inference latency. Without optimization, temporal attention layers recompute attention over the entire frame history at every step. We modify the WorldGym \cite{quevedo2025worldgymworldmodelenvironment} architecture for fast inference by introducing Key-Value (KV) caching \citep{pope2023efficiently} as described in Algorithm~\ref{alg:wm_kv_cache}. This change substantially reduces per-step latency especially with long context, enabling long-horizon RL training that would otherwise be infeasible. In practice, we observe a 10× reduction in rollout time when generating 20 trajectories in parallel over 40 frames.

\begin{algorithm}[t]
\caption{WorldGym inference with temporal KV caching.}
\label{alg:wm_kv_cache}
\begin{algorithmic}
\STATE \textbf{Init:}
\STATE \hspace{1em} $x \leftarrow x_0,\ \texttt{curr\_frame} \leftarrow 0,\ 
\texttt{chunk}=1,\ \texttt{max\_frames}$ \algcomment{key variables} %
\STATE \hspace{1em} $\mathrm{wm} \leftarrow \mathrm{WorldModel}(\texttt{checkpoint},\ \texttt{use\_kv\_cache}=\texttt{True})$
\STATE \hspace{1em} $\mathrm{wm.reset}(x_0)$ \algcomment{clears KV cache}
\\
\FOR{\textbf{each} action $a_t$}
  \STATE $\mathrm{wm.clear\_kv\_cache}()$ \algcomment{per-chunk cache}
  \STATE append noisy latent to history
  \STATE $\texttt{start} \leftarrow \max(0,\ \texttt{curr\_frame} + \texttt{chunk} - \texttt{max\_frames})$
  \FOR{\textbf{each} diffusion step}
    \STATE $\texttt{cache\_idx} \leftarrow \texttt{boundary\_of\_clean\_frames}(\texttt{start})$ \algcomment{$t{=}0$ window}
    \STATE $v_{\text{cond}} \leftarrow \mathrm{DiT}(x, t, a_t, \texttt{cache\_idx}, \texttt{start}, \texttt{cache}=\texttt{``cond''})$
    \STATE $v_{\text{null}} \leftarrow \mathrm{DiT}(x, t, \texttt{null}(a_t), \texttt{cache\_idx}, \texttt{start}, \texttt{cache}=\texttt{``null''})$
    \STATE $v \leftarrow \mathrm{cfg}(v_{\text{cond}}, v_{\text{null}})$
    \STATE $x \leftarrow \mathrm{ddim\_update}(x, v, \texttt{maybe\_cache\_suffix})$
  \ENDFOR
  \STATE decode latest clean frame(s)
  \STATE $\texttt{curr\_frame} \leftarrow \texttt{curr\_frame} + \texttt{chunk}$
\ENDFOR
\end{algorithmic}
\end{algorithm}

\subsection{OpenVLA-OFT as a Base Model}
For all experiments, we use OpenVLA-OFT \citep{kim2025fine} as our base model. OpenVLA-OFT builds on OpenVLA \cite{kim2024openvla} by providing a finetuning recipe designed to improve training efficiency and performance. In particular, it supports parallel decoding, action chunking, and a continuous action representation, resulting in faster and more performant inference and learning. We finetune OpenVLA-OFT starting from the OpenVLA checkpoint pretrained on the Open X-Embodiment dataset \cite{vuong2023open}, using the Bridge Dataset V2 \cite{walke2023bridgedata}.

\newpage
\subsection{Details of VLM as Reward}
\subsubsection{Prompt for VLM as Reward}
Prompt GPT-4o~\citep{hurst2024gpt} as Reward $\hat R$.
\begin{tcolorbox}[
  colback=gray!12,,
  colframe=gray!55,
  boxrule=0.5pt,
  arc=2pt,
  left=6pt,
  right=6pt,
  top=6pt,
  bottom=6pt
]
\small\ttfamily
Here is a sequence of frames from a robot policy which has been rolled out in a
video-generation-based world model.
I need your help determining whether the policy is successful. How successfully
does the robot complete the following task?

Instruction: \\
\{instruction\} \\
\{rubric.strip()\} \\
Provide brief reasoning (2--3 sentences). Then output EXACTLY one final line: \\
Final Score: X \\
Where X is \{ 'one of '0 or 1' \}. \\
No extra numbers after that line. \\

Note: Since this video was generated by a video prediction model (conditioned on
robot actions), it may contain some artifacts due to the video model capacity.
\end{tcolorbox}

The rubric is:

\begin{tcolorbox}[
  colback=gray!12,,
  colframe=gray!55,
  boxrule=0.5pt,
  arc=2pt,
  left=6pt,
  right=6pt,
  top=6pt,
  bottom=6pt
]
\small\ttfamily
Score rubric: \\
0 = Failure: instruction "\{instruction\}" not completed. \\
1 = Success: instruction completed.
\end{tcolorbox}
For each policy evaluation, we perform 5 independent rollouts in the world model.
To reduce token usage and mitigate redundancy in adjacent frames, each rollout video
is temporally downsampled with a stride of 3 before being sent to the VLM for scoring.

\newpage
\subsection{Details of Hyperparameters for World Model and RL Training}\label{app:hyperparams}
\textbf{World Model Implementation Details.} WorldGym~\citep{quevedo2025worldgymworldmodelenvironment} serves as the world-model-based simulator throughout our experiments. It encodes $256\times256$ RGB image frames into a latent space using a pretrained VAE from Stable Diffusion 3~\citep{esser2024scaling}. The underlying world model in WorldGym is a 16-layer transformer with a hidden dimension of 1024 and 16 attention heads.
Unless otherwise specified, we follow the default WorldGym configuration for visual encoding, world model architecture, and rollout settings.We use a world model initialized from a pretrained checkpoint trained on the Open-X Embodiment dataset~\citep{vuong2023open}.

WorldGym represents actions as 10-dimensional vectors. In our experiments, we use the first 7 dimensions to match the OpenVLA~\citep{kim2024openvla} action space, consisting of a 6-dimensional end-effector pose and a binary gripper state. WorldGym is trained with a fixed context length of 20 frames; during longer rollouts, it conditions on a sliding window of the most 20 recent frames.

\begin{table}[ht]
\centering
\begin{tabular}{ll}
\toprule
\textbf{Hyperparameter} & \textbf{Value}  \\
\midrule
Total parameters & 609 M \\
Image Resolution & 256$\times$256 \\
DiT Patch Size & 2 \\
Input Channels & 16 \\
Hidden Size & 1024 \\
Layers & 16 \\
Attention Heads & 16 \\
MLP Ratio & 4 \\
Optimizer & AdamW (weight decay = $0.002$, $\beta_1 = 0.9, \beta_2 = 0.99$) \\
Learning rate & 8e-5 \\
Batch size & 16 \\
Action dimension & 10 \\
Training hardware & 2xA100 80GB \\
Training steps & 580k \\
Diffusion noise schedule & sigmoid \\
Sampling timesteps & 10 \\
Prediction target & $v$ \\
\bottomrule
\end{tabular}
\caption{Hyperparameters for training \method's video world model.}
\label{tab:hyper}
\end{table}

\newpage
\textbf{RL Training Implementation Details.} 
We train policies using Group Relative Policy Optimization (GRPO)~\citep{shao2024deepseekmath} within WorldGym~\citep{quevedo2025worldgymworldmodelenvironment}. For each task, training is initialized from a single observation frame and a language instruction. Policies are rolled out for up to 40 steps in the world model, and a binary task completion reward is assigned to each rollout using GPT-4o~\citep{hurst2024gpt}. Actions are generated in chunks of length 5, and policy updates follow a clipped policy gradient objective. All RL trainings are conducted entirely within WorldGym.

\begin{table}[ht]
\centering
\begin{tabular}{ll}
\toprule
\textbf{Hyperparameter} & \textbf{Value}  \\
\midrule
RL algorithm & GRPO \\
World model environment & WorldGym \\
Base policy & OpenVLA-OFT \\
Max rollout length & 40 steps \\
Action chunk length & 5 \\
Reward type & Binary task completion \\
Reward source & GPT-4o \\
Group size & 8 \\
Batch size & 20 \\
Learning rate & $5 \times 10^{-6}$ \\
Clip ratio & $\epsilon_{\text{high}}=0.28,\ \epsilon_{\text{low}}=0.2$ \\
Temperature & 1.6 \\
Entropy coefficient & 0.0 \\
Gradient clipping & 1.0 \\
Optimizer & AdamW \\
Training hardware & 4 $\times$ NVIDIA H200 (140GB) \\
Training duration & 1--2 days \\
\bottomrule
\end{tabular}
\caption{Hyperparameters for RL training in WorldGymnast.}
\label{tab:rl_hyper}
\end{table}

\newpage
\section{Baselines}\label{app:baselines}
\subsection{Supervised Fine-tuning}
Supervised fine-tuning is an alternative for reinforcement learning for policy improvement due to its sample efficiency. As our first baseline, we follow the recipe of OpenVLA-OFT ~\cite{kim2025fine} to fine-tune a OpenVLA-7B model ~\cite{kim2024openvla} on expert trajectories from the Bridge V2 dataset ~\cite{walke2023bridgedata}. 

Methods like Ctrl-World ~\cite{guo2025ctrl} explore iterative supervised fine-tuning with synthetic data generated from the world model. Following their approach, we pick the same held-out scenarios used in RL training, and roll out the base SFT policy with our world model pretrained on OpenX Embodiment dataset ~\cite{vuong2023open}. To ensure fair comparison, the roll-out steps for iterative SFT is the same as total roll-out steps during RL fine-tuning. We then prompt a VLM to filter for successful trajectories, which are mixed with Bridge V2 dataset for the next iteration of fine-tuning, with a $1:1$ sampling rate.

\subsection{SIMPLER}\label{app:simpler} 
To compare \method with simulator-based RL, we select the SIMPLER simulator \cite{li2024evaluating} which provides a digital twin of the Bridge robot setup. SIMPLER includes the following tasks: 1) \textit{put spoon on table cloth}, 2) \textit{put carrot on plate}, 3) \textit{stack green cube on yellow cube}, 4) \textit{put eggplant in basket}, 5) \textit{put eggplant in sink}, 6) \textit{open drawer} and 7) \textit{close drawer}. The last 4 tasks are digital twins of AutoEval setup \cite{zhou2025autoeval}. 

The base SFT policy showed poor performance on SIMPLER tasks, resulting in low reward variance and unstable training with GRPO \cite{shao2024deepseekmath}. To address this, we define trajectory reward as sum of step rewards. Each step is assigned partial credit using a dense reward scheme: a reward of 0.1 was awarded for each of the following conditions: \textit{is source object grasped}, \textit{is grasped}, \textit{consecutive grasp}, \textit{lifted object significantly} and \textit{lifted object}. Success earned reward of value 1.

\newpage
\section{Datasets}\label{app:data}
\subsection{OpenVLA Evaluation Task Set}
We evaluate different methods on a curated set of tabletop manipulation tasks originally introduced in OpenVLA~\citep{kim2024openvla}. 
This task suite is designed to assess policy generalization across multiple axes, including visual, motion, physical, and semantic variations, as well as language grounding in scenes with multiple objects. 
For our RL training, the task suite is split into an 80/20 train–evaluation split, where both splits share the same set of tasks but differ in initial frames.

\begin{table}[ht]
\centering
\begin{tabular}{ll}
\toprule
\textbf{Tasks} & \textbf{Generalization Types}  \\
\midrule
Put Eggplant into Pot (Easy Version) & Visual Generalization \\
Put Eggplant into Pot & Visual Generalization \\
Put Cup from Counter into Sink & Visual Generalization \\
Put Eggplant into Pot (w/ Clutter) & Visual Generalization \\
Put Yellow Corn on Pink Plate & Visual Generalization \\
Lift Eggplant & Motion Generalization \\
Put Carrot on Plate (w/ Height Change) & Motion Generalization \\
Put Carrot on Plate & Physical Generalization \\
Flip Pot Upright & Physical Generalization \\
Lift AAA Battery & Physical Generalization \\
Move Skull into Drying Rack & Semantic Generalization \\
Lift White Tape & Semantic Generalization \\
Take Purple Grapes out of Pot & Semantic Generalization \\
Stack Blue Cup on Pink Cup & Semantic Generalization \\
Put {Eggplant, Red Bottle} into Pot & Language Grounding \\
Lift {Cheese, Red Chili Pepper} & Language Grounding \\
Put {Blue Cup, Pink Cup} on Plate & Language Grounding \\
\bottomrule
\end{tabular}
\caption{Tasks and corresponding Generalization Types for OpenVLA Evaluation Task Set.}
\label{tab:openvla-eval}
\end{table}

\subsection{Visual Distractors}
To evaluate robustness to visual perturbations, we construct a distractor-augmented training distribution by modifying initial frames from the OpenVLA evaluation task set~\citep{kim2024openvla}. Using Nano Banana~\citep{SharonEtAl2025GeminiImageEditing}, we insert visually diverse but task-irrelevant objects into the scene while keeping the underlying task, goal specification, and robot configuration unchanged. The distractor dataset is constructed from the training split of OpenVLA evaluation task set. Specifically, 90\% of training episodes use the original initial frames, while 10\% are initialized from distractor-augmented frames. 

For evaluation, we measure performance both on a held-out set of distractor-augmented initial frames and on the original OpenVLA evaluation tasks in WorldGym~\citep{quevedo2025worldgymworldmodelenvironment}, to assess robustness under visual perturbations. Qualitative rollout examples under visual distractors are included in Figure~\ref{fig:distractor-qualitative}

Prompt for Nano Banana to edit initial frames.
\begin{tcolorbox}[
  colback=gray!12,,
  colframe=gray!55,
  boxrule=0.5pt,
  arc=2pt,
  left=6pt,
  right=6pt,
  top=6pt,
  bottom=6pt
]
\small\ttfamily
This is an initial observation for a robotics task \texttt{[TASK\_INSTRUCTION]}.
Modify this image by adding distraction objects in the scene in a natural way
without moving or changing any objects in the original scene.

Requirements:
\begin{itemize}
  \item The robotic arm should be visible.
  \item With the distractors, the task \texttt{[TASK\_INSTRUCTION]} should remain achievable.
\end{itemize}
\end{tcolorbox}

\subsection{Out-of-Distribution Languages}
To evaluate robustness to language distribution shift, we construct an out-of-distribution (OOD) language training distribution by modifying task instructions for a subset of OpenVLA tasks~\citep{kim2024openvla}. 
The underlying scenes and objects remain unchanged; only the natural language instructions are altered to describe new interactions.

The OOD language dataset consists of four newly defined language variants applied to existing OpenVLA tasks:
\textbf{(a)}: \emph{move pot with grapes into the drying rack}, 
\textbf{(b)}: \emph{pick up the pot from the drying rack and place it outside the sink on the counter}, 
\textbf{(c)}: \emph{put plate on drying rack}, and 
\textbf{(d)}: \emph{put yellow corn in red cup}. 
During RL training, we combine this OOD language data with the training split of OpenVLA evaluation task set using a 50/50 mixture over initial frames between the original and OOD language instructions. Qualitative rollout examples under OOD language instructions are included in Figure~\ref{fig:oodlang}.

\subsection{Evaluating Diverse Settings \method Offers}

\subsection{Scaled Task Set}
To study the effect of increasing task diversity during RL training, we construct a scaled training distribution by augmenting the training split of the OpenVLA evaluation task set with five additional manipulation tasks randomly selected from the Bridge Dataset V2~\citep{walke2023bridgedata}:
\textbf{(a)}: \emph{close cabinet}, 
\textbf{(b)}: \emph{flip orange pot upright in sink}, 
\textbf{(c)}: \emph{fold the cloth from right to left}, 
\textbf{(d)}: \emph{opened the drawer}, and
\textbf{(e)}: \emph{take spatula off plate sink}. 
These additional tasks are not present in the OpenVLA evaluation task suite and are used only during RL training, while the training procedure remains unchanged.

\subsection{Real-Robot Evaluation Dataset}
To evaluate real-world transfer, we report performance on AutoEval~\citep{zhou2025autoeval}, an automated real-robot evaluation framework built on tasks from the Bridge Dataset V2 distribution. 
In our experiments, we evaluate policies on three manipulation tasks supported by AutoEval across two scenes using a WidowX robot arm: two drawer manipulation tasks (\emph{open the drawer}, \emph{close the drawer}) and one pick-and-place tasks in the sink scene (\emph{put the eggplant in the blue sink}). We do not evaluate on the cloth manipulation task supported by AutoEval. Qualitative rollout examples under scaled task set are included in Fig~\ref{fig:autoeval}.

\newpage
\section{Evaluation}
\subsection{WorldGym Evaluation}
We evaluate learned policies in WorldGym~\citep{quevedo2025worldgymworldmodelenvironment}, a world-model-based simulator designed to approximate real-robot execution.

\paragraph{Rollout protocol.} For each task, the policy is rolled out in WorldGym for up to 40 steps starting from a given initial frame and language instruction.  Each rollout produces a sequence of predicted future frames, which are used to assess task completion.

\paragraph{Success evaluation.} Following the WorldGym evaluation protocol, task success is determined using a vision-language model (VLM) that compares the rollout outcome against the task instruction. To reduce variance in VLM-based scoring, we obtain 5 independent VLM judgments per rollout and assign a binary success label by majority vote.
\subsection{Autoeval Evaluation}

\begin{figure}
    \centering
    \includegraphics[width=\linewidth]{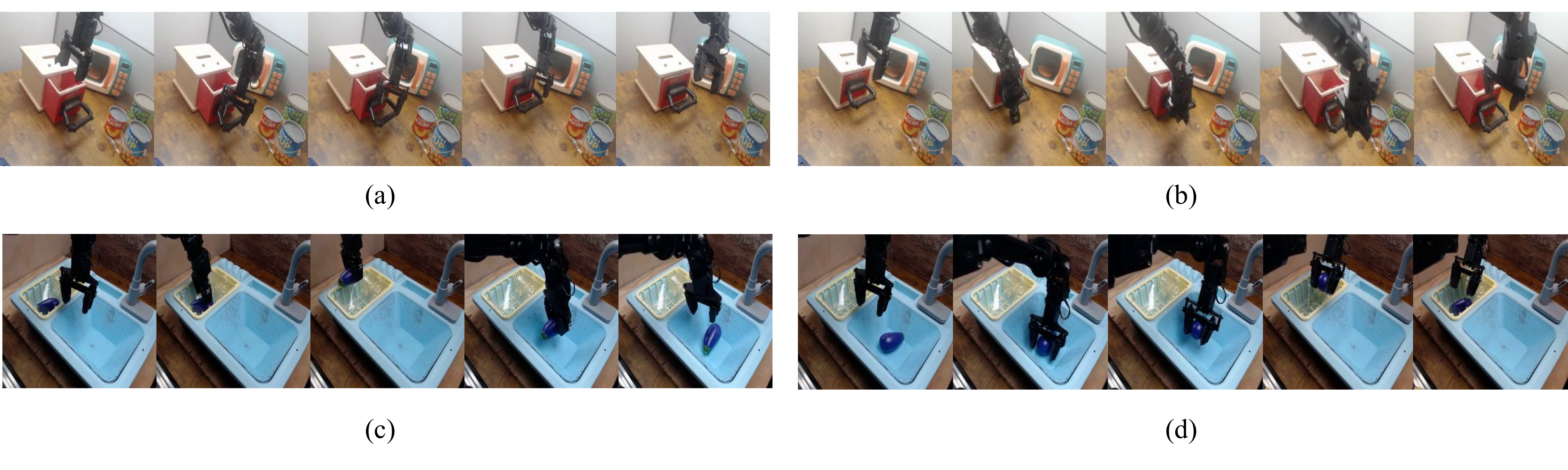}
    \caption{\textbf{Qualitative evaluation of policy rollouts in AutoEval.} \textbf{(a):} \textit{close the drawer}; \textbf{(b):} \textit{open the drawer}; \textbf{(c):} \textit{put the eggplant in the blue sink}; \textbf{(d):} \textit{put the eggplant in the yellow basket}.\ }
    \label{fig:autoeval}
\end{figure}

AutoEval ~\cite{zhou2025autoeval} provides a convenient and reliable platform for policy evaluation on real world robots. There are two setups available, each equipped with a WidowX robotic arm (Figure \ref{fig:autoeval}). Each setup can run two tasks which perform the opposite operation to one another (e.g. open and close drawer). When evaluating a policy on a task, there is a reset policy to reset the conditions, allowing the policy being evaluated to conduct next trial.

\newpage

\begin{figure}
    \centering
    \includegraphics[width=\linewidth]{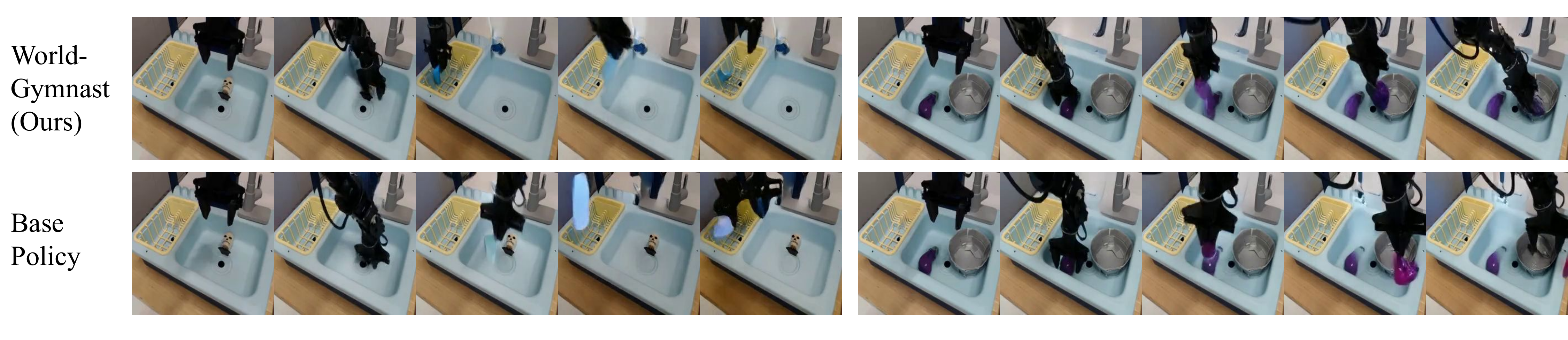}
    \caption{\textbf{Qualitative evaluation of policy rollouts in WorldGym.} We compare the \method policy fine-tuned with RL and the base policy before performing RL. \textbf{Left:} \textit{lift skull}; \textbf{Right:} \textit{put eggplant in pot}.}
    \label{fig:base_wm}
\end{figure}

\begin{figure}
    \centering
    \includegraphics[width=\linewidth]{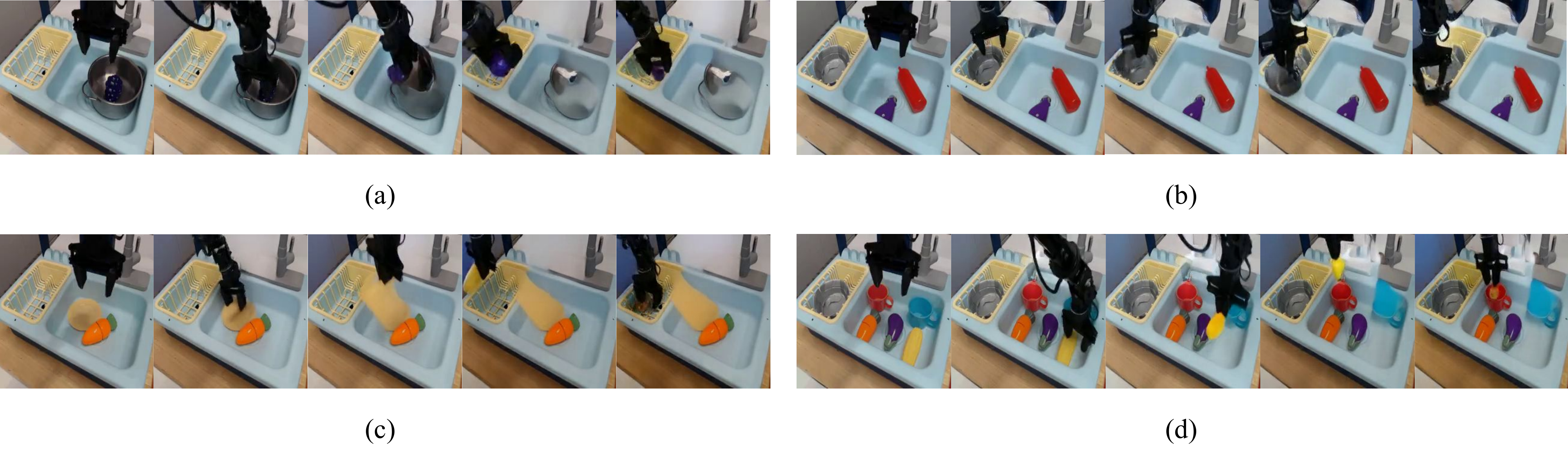}
    \caption{\textbf{Qualitative evaluation of policy rollouts with out-of-distribution language descriptions.} \textbf{(a):} \textit{move pot with grapes into the drying rack}; \textbf{(b):} \textit{pick up the pot from the drying rack and place it outside the sink on the counter}; \textbf{(c):} \textit{put plate on drying rack}; \textbf{(d):} \textit{put yellow corn in red cup}.\ }
    \label{fig:oodlang}
\end{figure}

\begin{figure}
    \centering
    \includegraphics[width=\linewidth]{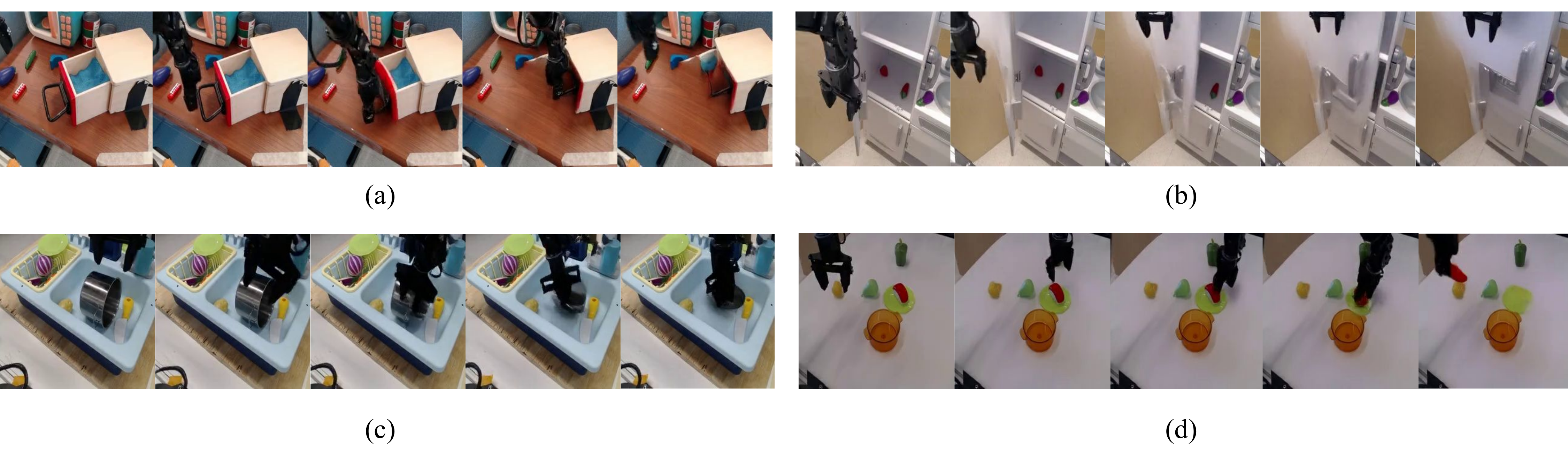}
    \caption{\textbf{Qualitative evaluation of policy rollouts given novel initial image frames.} \textbf{(a):} \textit{close the drawer}; \textbf{(b):} \textit{close fridge}; \textbf{(c):} \textit{flip pot upright which is in sink}; \textbf{(d):} \textit{take sushi off plate}.}
    \label{fig:novel_frames}
\end{figure}


\begin{figure}
    \centering
    \includegraphics[width=\linewidth]{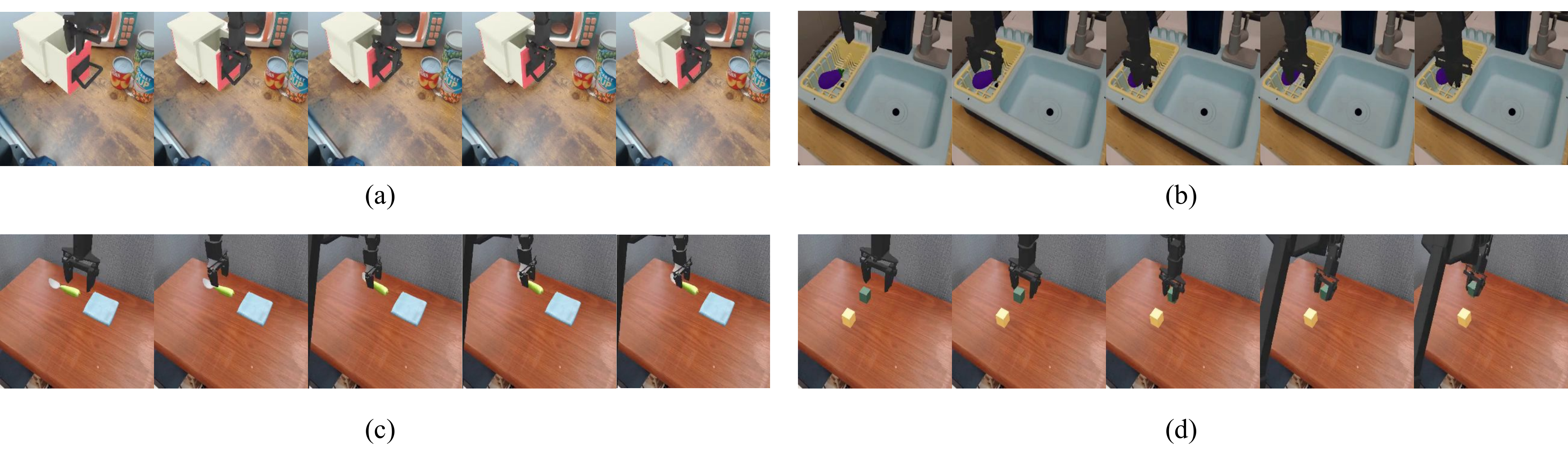}
    \caption{\textbf{Qualitative evaluation of the policy trained in the SIMPLER simulator.} \textbf{(a):} \textit{close drawer}; \textbf{(b):} \textit{put the eggplant in sink}; \textbf{(c):} \textit{put spoon on table cloth}; \textbf{(d):} \textit{stack green cube on yellow cube}.}
    \label{fig:simpler}
\end{figure}

\begin{figure}
    \centering
    \includegraphics[width=\linewidth]{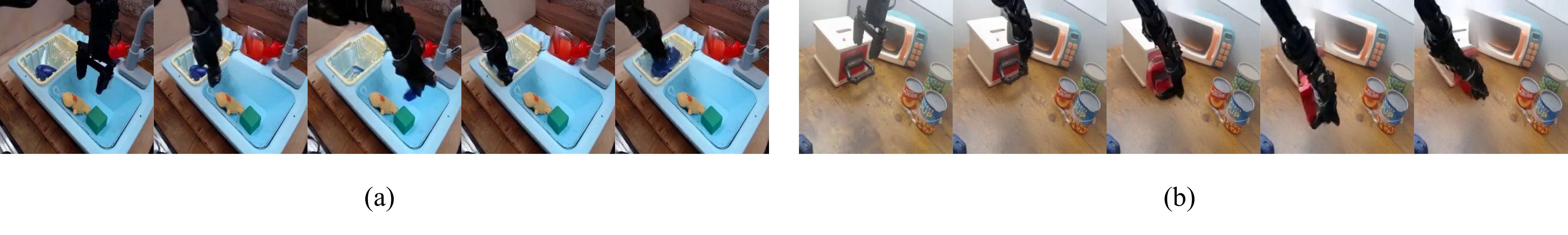}
    \caption{\textbf{Qualitative evaluation of policy rollouts in WorldGym with frames from AutoEval.} \textbf{(a):} \textit{put the eggplant in the
    blue sink}; \textbf{(b):} \textit{open the drawer}. WorldGym was not trained on these image observations from AutoEval.}
    \label{fig:test_time}
\end{figure}

\section{Qualitative examples under difference scenarios}

In this section, we provide additional qualitative results and visualization to supplement our reported results.

\paragraph{Rollout visualization.} In Figure \ref{fig:base_wm}, we visualize a few rollout samples in the learned world model WorldGym. We also present a comparison between the base policy and the RL-finetuned policy from \method, illustrating the effectiveness of RL with a world model. Figure \ref{fig:autoeval} shows some examples of the AutoEval evaluation settings. On the other hand, Figure \ref{fig:test_time} displays how WorldGym performs with the slightly out-of-distribution AutoEval image frames.

\paragraph{Diverse training scenarios for \method.} We provide additional qualitative results on the discussed diverse training settings enabled by \method in Figure \ref{fig:oodlang}. Figure \ref{fig:novel_frames} shows the policy behavior when trained on unseen scenes in the world model without expert data.

\paragraph{Comparison with SIMPLER.} Here we show in Figure \ref{fig:simpler} additional qualitative evaluations of a policy trained with RL in the SIMPLER simulator, as discussed in Section \ref{sec:exp-simpler-sft}. Training in such a simulator proves to be less desirable for real-life deployment.




\end{document}

%% file: introduction.tex
\section{Introduction}

Robots that learn by trial and error in the real world face an inherent constraint: physical interaction is expensive~\citep{kormushev2013reinforcement}. Every policy update that depends on executing actions on hardware consumes operator time, risks wear-and-tear, and compounds safety concerns~\citep{brunke2022safe}, especially for manipulation, where failures are frequent early in learning. This cost creates a fundamental bottleneck for scaling robot learning from interaction. As a result, many real-robot systems rely on alternatives that reduce or replace on-robot exploration~\citep{matas2018sim,schaal1999imitation}.

One alternative is supervised learning (SFT) from expert demonstrations, where a robot is trained to imitate trajectories collected by teleoperation or scripted controllers~\citep{ross2011reduction}. However, demonstration data tend to cover only a narrow slice of the long tail situations~\citep{hu2024data}, and rarely expose the robot to the kinds of compounding errors and recovery behaviors needed for robust deployment~\citep{lu2022challenges}. The second alternative is reinforcement learning (RL) in a software-based simulator~\citep{zhao2020sim}. However, software simulators are costly to create for every new scenarios. Furthermore, they often suffer from the sim-to-real gap where visual features differ from real-world images~\citep{salvato2021crossing}.

Recent work have shown that world models learned from real-robot data can approximate real-robot execution outcomes~\citep{yang2023learning,quevedo2025worldgymworldmodelenvironment,guo2025ctrl,tseng2025scalable,li2025worldeval}. These models aim to predict how the visual world evolves under the robot’s actions, effectively serving as an action-conditioned video simulator learned from real-world data. Compared to software simulators, video-based world models hold the promise of closing the visual gap and generalizing to a novel initial frame. However, it is unclear whether video world model offers more realistic physics than traditional simulators due to hallucinations~\citep{yang2024video}. While evaluating physical realism is difficult, we instead tackle the end-to-end problem: does training robot policies inside a learned world model result in better real-robot performance than SFT or RL in a traditional simulator?

\begin{figure*}[t]
    \centering
    \includegraphics[width=\linewidth]{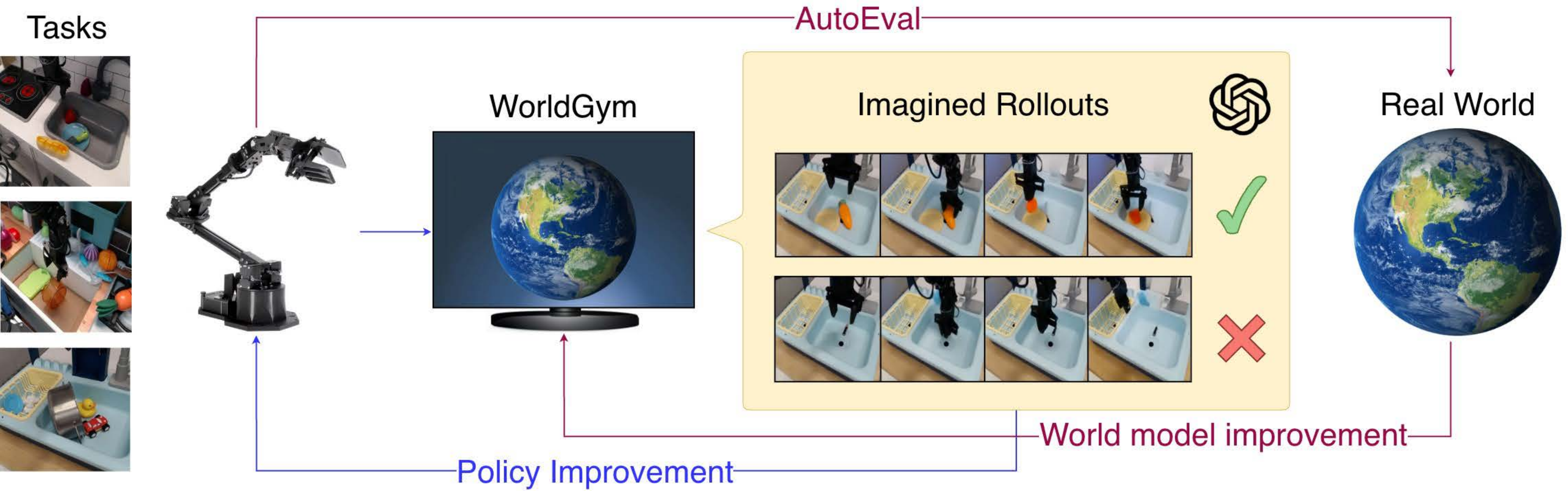}
    \caption{\textbf{Overview of \method.} The policy is trained on tasks specified by an initial frame and language instruction. During training, the policy outputs actions which are then passed to the world model (WorldGym \cite{quevedo2025worldgymworldmodelenvironment}) which generates imagined rollouts. These rollouts are then passed to a VLM which returns a binary task completion reward. This reward is used to update the policy. Once trained, we evaluate the policy on real robots using the AutoEval \citep{zhou2025autoeval} setup. The resulting real world rollouts (frame-action sequences) from AutoEval can be further used to improve the world model on the particular environment.}
    \label{fig:intro-figure}
\end{figure*}

In this paper, we explore this question through the lens of large vision-language-action (VLA) policies that map images and language instructions to robot actions~\citep{brohan2022rt,kim2024openvla,black2410pi0}. Specifically, we propose \method, a training framework that performs RL fine-tuning a VLA policy using a world model (Figure \ref{fig:intro-figure}). Concretely, \method uses the action-conditioned video generation model similar to \citet{quevedo2025worldgymworldmodelenvironment} as its world model, enabling the policy to generate imagined rollouts conditioned on action sequences sampled from the VLA and uses a vision-language model (VLM) to compute rewards from predicted video frames. The resulting rewards are used to perform policy gradient updates to the VLA policy. More importantly, \method opens up many intriguing possibilities of RL training with a world model, including (i) RL training from an arbitrary image frame, (ii) test-time training on a novel initial frame, and (iii) online iterative world model and policy improvement.

We evaluate \method on the Bridge robot platform through AutoEval~\citep{zhou2025autoeval}, an automated real-robot evaluation platform open to public. Across a suite of manipulation tasks from AutoEval, we show \method substantially outperforms SFT using the original Bridge data~\citep{walke2023bridgedata} and RL in SIMPLER~\citep{li2024evaluating}, a software simulator for Bridge created through real-to-sim techniques. Furthermore, since the world model only requires a single initial frame to perform rollouts, we demonstrate intriguing usage of the world model including training on novel language instructions and initial frames injected with distractor objects, test-time training from a real-robot frame, and iterative world model and policy improvement, all of which positively contribute to improved real-robot performance.

%% file: preliminaries.tex
\section{Preliminaries}

In this section, we define notations and review model-based RL. We then discuss how foundation world models and vision language models can serve as general dynamics and reward models under the model-based RL formulation.

\paragraph{Markov Decision Process.} We consider a multi-task, finite-horizon, partially observable Markov Decision Process (POMDP)~\citep{puterman2014markov,kaelbling1995partially}, specified by $\mathcal{M} = (S, A, O, G, R, T, \mathcal{E}, H)$, which consists of state, action, observation, and task spaces, reward, transition, and emission functions, and horizon length. A policy $\pi$ interacts with the environment for a task starting from an initial state $g, s_0\sim G, o_0\sim\mathcal{E}(s_0)$, producing a distribution $\pi(\cdot|o_t, g)$ over $A$ from which an action $a_t$ is sampled and applied to the environment at each step $t\in [0, H]$. The environment produces a scalar reward $r_t = R(s_t, g)$, and transitions to a new state $s_{t+1}\sim T(s_t, a_t)$ and emits a new observation $o_{t+1}\sim\mathcal{E}(s_{t+1})$. 
    
The value of a policy $\pi$ can be defined as the total expected future reward:
\begin{align}
\rho(\pi) =& \mathbb{E}[R(s_H, g)|s_0, g\sim G, o_t\sim\mathcal{E}(s_t), a_t\sim\pi(o_t, g),\nonumber\\  
&s_{t+1}\sim T(s_t, a_t)\,\,\,\, \forall t\in[0, H]].
\end{align}

\paragraph{Model-Based RL with Foundation Models.} RL~\citep{sutton1998reinforcement} aims to maximize $\rho(\pi)$ through trial-and-error interactions between the policy and the environment. Model-based RL~\citep{doya2002multiple} considers the setting where $T$ and $R$ are unknown and need to be estimated from samples from the environment, which can be an offline dataset logged from previous interactions $D=\{\tau_i = g, s_0, o_0, a_0, ..., s_H, o_H, r_H\}$. Motivated by characteristics of a real-world system such as image based observations and high control frequencies, the learned model $\hat T(\cdot|\mathbf{o}, \mathbf{a})$ can often take a sequence of previous image observations and a sequence of next actions. After $\hat T$ and $\hat R$ are estimated from data, a policy can perform rollout in the learned model 
\begin{align}
\hat\rho(\pi) =& \mathbb{E}[\hat R([o_0, ..., o_H], g)|s_0, g\sim G, \mathbf{a}\sim\pi(\mathbf{o}, g), \nonumber \\& \mathbf{o'}\sim \hat T(\mathbf{o}, \mathbf{a}), \mathbf{o} = \mathbf{o'}].\label{eq:maxret}
\end{align}
Recent work has shown that $\hat T$ can be parametrized using an action-conditioned video generation model (world model) while $\hat R$ can be parametrized using a vision-language model (VLM).

Policy gradient methods~\citep{williams1992simple} estimates the gradient of Equation \ref{eq:maxret} with respect to the policy $\pi$, and maximizes $\rho(\pi)$ directly via gradient ascent. The most commonly used gradient estimator has the form
\begin{equation}
    \nabla_\theta\rho(\pi_\theta) = E_{\tau\sim \pi, T}\left[\textstyle{\displaystyle}\sum_{t=0}^H \gamma^t\nabla_\theta\log\pi_\theta(o_t, g)\hat{A}(o_t, a_t)\right],\label{eq:pg}
\end{equation}
where $\hat{A}$ is some advantage function that can be separately estimated via Monte-Carlo returns from $\pi, T, R$~\citep{williams1992simple}. With model-based policy gradient, these advantages can be estimated from Monte-Carlo samples from $\pi, \hat T, \hat R$.

%% file: method.tex
\section{RL with a World Model}

In this section, we describe the RL algorithm \method uses in Section~\ref{sec:method-algo}. We then describe emerging training scenarios such as training on out-of-distribution (OOD) language and image in Section~\ref{sec:method-ood} and test-time training in Section~\ref{sec:method-test-time}. Lastly, we explain how \method can be combined with classical algorithm such as Dyna~\cite{sutton1991dyna} to do online iterative world model and policy improvements.

\subsection{Model-Based GRPO with World Model Rollouts}\label{sec:method-algo}

To optimize the policy $\pi_\theta$ from Equation~\eqref{eq:pg}, \method uses the learned world model $\hat{T}$ from \citet{quevedo2025worldgymworldmodelenvironment}. We adopt Group Relative Policy Optimization (GRPO)~\citep{shao2024deepseekmath}, a policy gradient algorithm that  estimates $\hat A$ using group-based score normalization.

For a given task instruction $g$ and an initial observation $o_0$, we generate a group of $K$ independent trajectories $\{\tau_1, \dots, \tau_K\}$ by rolling out the policy $\pi_\theta$ in the world model $\hat{T}$. Specifically, for the $k$-th trajectory, the policy samples an action $a_{t,k} \sim \pi_\theta(\cdot|o_{t,k}, g)$, and the world model predicts the next observation $o_{t+1,k} \sim \hat{T}(o_{t,k}, a_{t,k})$. This process repeats until the horizon $H$ is reached, yielding a trajectory $\tau_k = (o_{0,k}, a_{0,k}, \dots, o_{H,k})$. Once the rollouts are complete, we employ a VLM $\hat{R}$ to assign a binary task completion reward to each trajectory $r_k = \hat{R}(\tau_k, g)$. To compute the advantages, we treat the group of $K$ outputs as a baseline. We compute the mean and standard deviation of the rewards within the group:
\begin{equation}
    \mu = \frac{1}{K}\sum_{k=1}^K r_k, \quad \sigma = \sqrt{\frac{1}{K-1}\sum_{k=1}^K (r_k - \mu)^2}.
\end{equation}
The advantage for the $k$-th trajectory is then calculated via normalization:
\begin{equation}
    \hat{A}_k = \frac{r_k - \mu}{\sigma + \epsilon},
\end{equation}
where $\epsilon$ is a small constant for numerical stability. We assign the trajectory-level advantage to every time step $t$ within that trajectory. That is, $\hat{A}_{t,k} = \hat{A}_k$ for all $t \in [0, H-1]$. Finally, we optimize the policy $\pi_\theta$ using a PPO-style objective clipped based on the computed advantages. The loss function is defined as:
\begin{align}
    \mathcal{J}(\theta) &= \mathbb{E}_{g, o_0 \sim \mathcal{D}} \bigg[ \frac{1}{K} \sum_{k=1}^K \frac{1}{H}\sum_{t=0}^{H-1} \nonumber\\
    &\min \left( r_{t,k}(\theta) \hat{A}_k, \text{clip}(r_{t,k}(\theta), 1-\epsilon_{low}, 1+\epsilon_{high}) \hat{A}_k \right) \bigg],
\end{align}
where $r_{t,k}(\theta) = \frac{\pi_\theta(a_{t,k}|o_{t,k}, g)}{\pi_{\theta_{old}}(a_{t,k}|o_{t,k}, g)}$ denotes the probability ratio.

Following the successful training setup of VLA training using RL in \citet{li2025simplevla}, we employ some of their techniques: 1) discarding the KL penalty term, 2) dynamic sampling to filter out groups with no variance in reward, 3) clipping higher in GRPO, and 4) using a higher temperature to sample actions during rollouts. These tricks helped stabilize training and improved exploration during training.

\subsection{Diverse Training Scenarios in the World Model}\label{sec:method-ood}
A world model pretrained on diverse datasets allows us to generate diverse training configurations (e.g., tasks and initial observations) using only images and langauge instructions. This provides greater flexibility than setting up software based simulations for each new configuration.
 We now explore an array of possibilities in training a policy in diverse configurations enabled by \method.

\paragraph{Training from Any Frame.} We can train the policy with RL using any frames that are close enough to the world model’s training distribution as the initial observation $o_0$, then rolling out the policy $\pi$, the world model $\hat T$, and the reward model $\hat R$ to provide learning signals on this initial configuration. This flexibility substantially increases the effective amount of training data available for RL in contrast to SFT which is bottlenecked by the amount of expert demonstrations. RL training from any frame also enables the policy to learn recovery behaviors, thereby improving the robustness of policies.

\paragraph{Training on Novel Language Instructions.} The training data can be further scaled by modifying the language instructions associated with the same initial frame. For instance, we can give a VLM an initial frame and ask for reasonable tasks for a robot to perform from that initial frame. We then give these reasonable tasks as language instructions to the VLA policy to evaluate the policy's performance on OOD language tasks and to further improve the policy on the OOD language tasks through RL. This enables the policy to be trained on new tasks and interact with objects previously present in the environment but not explicitly interacted with. Previous work in policy evaluation had shown that pretrained VLA policies often fail at following OOD langauge instructions~\citep{quevedo2025worldgymworldmodelenvironment}. We can overcome these limitations of VLAs with RL post-training in a world model.

\paragraph{Training with Distractions.} 
To improve policy's robustness to irrelevant visual clutter, we leverage image editing tools like Nano Banana \citep{SharonEtAl2025GeminiImageEditing} to synthesize additional objects as distractors in the input image frames. Training the policy on diverse distractor objects encourages the policy to be more robust when such distractor objects are present during actual deployment and to have better performance in cluttered scenes. This can bridge the gap between robots that work in demos and robots that can work in anyone's messy household.

\subsection{Test-Time Training from a Novel Frame}\label{sec:method-test-time} 
Because \method allows a policy to rollout from just an initial frame, when a novel frame is presented to a policy at test time, the policy can trade-off compute for improved policy performance by running RL training in the world model starting from the test frame. This allows rapid adaptation of the policy to novel scenes while avoiding the cost and risk of collecting real-world interaction data.


\subsection{Iterative World Model and Policy Improvement}\label{sec:method-dyna} 
When the visual observations encountered during policy rollouts  deviate too much from the original training distribution of the world model, directly rolling out the policy in the world model might lead to compounding modeling errors. Inspired by classical Dyna-style algorithms~\citep{sutton1991dyna}, \method allows an iterative training procedure in which the policy and world model are alternately refined. Specifically, the current policy can be rolled out (with inference-time scaling or test-time training using the world model as a reward function) to collect new environment interactions, which are then incorporated to further fine-tune the world model. The updated world model is subsequently used to generate improved imagined rollouts for policy optimization. This data flywheel enables the world model to progressively adapt to the policy-induced state distribution, while allowing the policy to benefit from increasingly accurate long-horizon predictions from the world model.

%% file: experiment.tex
\section{Experiments} \label{sec:exp}

We now evaluate the performance of policies trained in \method. We explain the experimental setup in Section~\ref{sec:exp-setup}, followed by comparisons to policies trained with software simulators and SFT in Section~\ref{sec:exp-simpler-sft}. We then demonstrate the capabilities of \method in supporting diverse training from images with distractors, novel language instructions, and scaling the number of RL tasks in Section~\ref{sec:exp-diverse}. Finally, we evaluate test-time scaling and iterative policy and world model improvement in Section~\ref{sec:exp-inference} and Section~\ref{sec:exp-dyna}.

\subsection{Experimental Setup}\label{sec:exp-setup}
\paragraph{Tasks and Pipeline.} We evaluate the efficacy of \method using the curated evaluation dataset used in \citet{kim2024openvla}. The dataset follows the BridgeData V2 \citep{walke2023bridgedata} setup with the WidowX robot and is designed to test policy generalization across visual, motion, physical, and semantic variations, as well as language grounding, over 17 tasks (Appendix~\ref{app:data}).

We further leverage the data scaling capabilities enabled by \method, such as image editing, language augmentation, and novel task setups, to improve generalization of the trained RL policy. To train with \method, a task just requires an initial frame and language instruction. During training, \method rolls out the policy for up to 40 steps in WorldGym and a binary task completion reward is assigned to the rollout by GPT-4o \citep{hurst2024gpt}. 

Once training is complete, we evaluate the performance of the policy on WorldGym \citep{quevedo2025worldgymworldmodelenvironment} to estimate real robot performance and ensure policy is safe for testing, following its default configuration unless otherwise specified (Appendix~\ref{app:hyperparams}). We finally run the policy on AutoEval \citep{zhou2025autoeval}, a real-robot evaluation framework that currently supports 4 tasks across 2 setups. AutoEval evaluates each policy–task pair over 10 trials; we repeat this evaluation 5 times to estimate the standard error.

\paragraph{Base Models.} Successful RL finetuning requires a reasonably competent initial policy. To this end, we use OpenVLA-OFT \citep{kim2025fine} as our base model. OpenVLA-OFT provides an optimized finetuning recipe to build on top of OpenVLA \citep{kim2024openvla} which was originally trained on Open X-Embodiment dataset \citep{vuong2023open}. We use the BridgeData V2 \citep{walke2023bridgedata} to finetune our base model. Following the idea from \cite{li2025simplevla}, we made several modifications to the official implementation of OpenVLA-OFT: 1) Disable the proprioception and secondary camera inputs to match the observation space used by our policy, 2) Use LLAMA-2 \citep{touvron2023llama} LM head as action head instead of the default L1 loss one to get action probabilities essential for RL. For WorldGym, we used a 600M parameter variant pretrained on Open X-Embodiment dataset.

\paragraph{Training Details.} For RL training, we use 4 NVIDIA H200 GPUs (140GB each) for full-parameter finetuning over 1–2 days. We use the following training parameters: learning rate $5\cdot10^{-6}$, group size $8$, size of training batch $20$, length of action chunk $5$, clip ratio ($\epsilon_{high} = 0.28$,  $\epsilon_{low} = 0.2$), temperature $1.6$. More detailed hyperparameter setup is available in Appendix~\ref{app:hyperparams}.

\begin{table}[t]
    \centering
    \caption{\textbf{Real-robot success rate from AutoEval}~\citep{zhou2025autoeval} of  \method compared to running RL in a software simulator SIMPLER~\citep{li2024evaluating}. RL with a world model significantly outperforms RL in a simulator in terms of real-robot success for 3 out of the 4 tasks.}
    \label{tab:rl-simpler}    
    \begin{tabular}{p{3cm}cc}
    \toprule
    Task & SIMPLER & \method \\
    \midrule
      Open the drawer   & $34\pm7\%$ & $58\pm4\%$ \\
      Close the drawer   & $74\pm5\%$ & $62\pm6\%$ \\
      Put the eggplant into the blue sink & $32\pm10\%$ & $72\pm10\%$ \\
      Put the eggplant into the yellow basket & $40\pm10\%$ & $78\pm2\%$ \\
    \bottomrule
    \end{tabular}
\end{table}

\subsection{Evaluating RL with \method}\label{sec:exp-simpler-sft}

\paragraph{Comparing \method to a Software Simulator.} We compare \method against traditional simulator-based RL. We select SIMPLER \citep{li2024evaluating}, a real-to-sim policy evaluation framework, as our baseline since it provides the closest simulator-based approximation to the Bridge robot and tasks used in our evaluation. We train on all available tasks in SIMPLER (Appendix \ref{app:simpler}) and further include digital twins for the AutoEval setup. When initializing RL from the base SFT policy, we observed low task completion rate in SIMPLER for all tasks except \textit{close the drawer}, although the policy often moved in the correct direction but failed to fully complete the task. As a result, using a binary completion reward led to collapsed reward variance, preventing effective policy updates. To overcome this problem, we define the reward for a rollout as the sum of partial rewards from each step. We share more details of reward design in Appendix \ref{app:simpler}. 

\method outperformed training with SIMPLER on all tasks except \textit{close the drawer}, as shown in Table~\ref{tab:rl-simpler}. For \textit{close the drawer} task, the base policy has already performed pretty well prior to RL. It is worth noting that the RL training set for \method does not include these tasks in the training set, whereas the SIMPLER baseline was also trained on these exact environment-task setups (the digital twins) and yet the policy exhibited poor transfer to real world. 

\begin{figure*}[t]
    \centering
    \includegraphics[width=\linewidth]{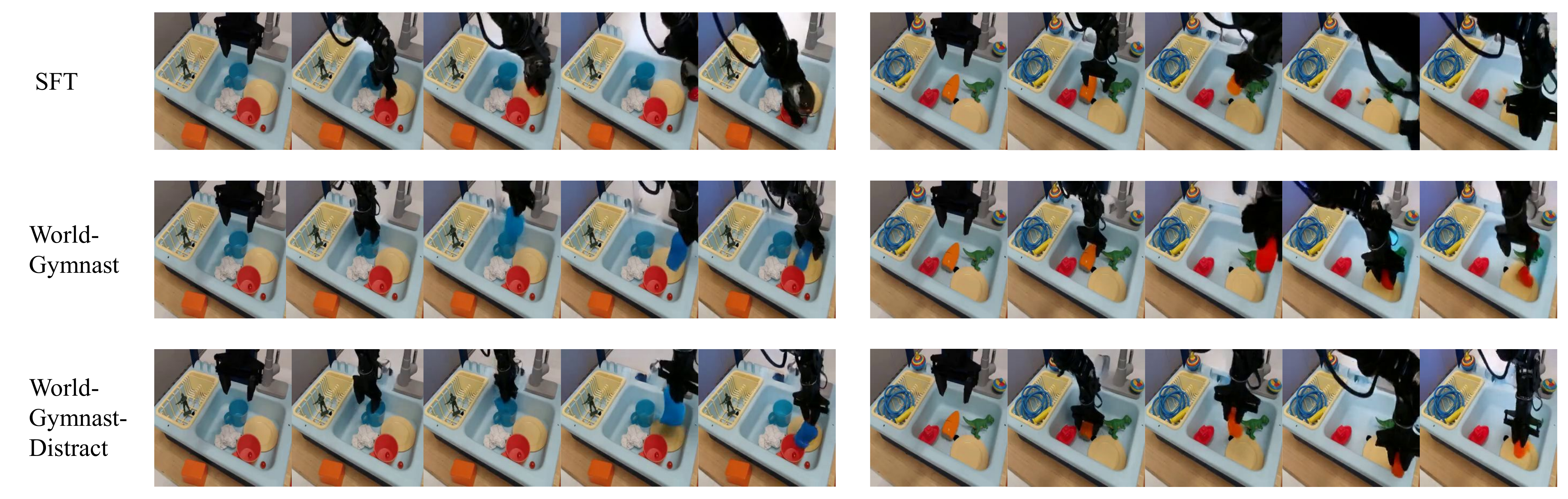}
    \caption{\textbf{Qualitative evaluation of policy rollouts in WorldGym with distractors.} We compare rollout quality among SFT, \method and \method-Distract under visual distractions. The task on the left is \textit{put blue cup on plate} and the SFT policy clearly picks up the wrong cup, while both \method variants are able to correctly execute the task. On the right task (\textit{put carrot on plate}), we can see SFT struggle again and seems to grab the dinosaur along with the carrot. Both \method variants are again successful but \method-Distract has better grasping and placing movements. It is worth noting that even with the visual artifacts introduced by the imperfect world model, the policies transfer effectively to the real robot setting.}
    \label{fig:distractor-qualitative}
\end{figure*}

\begin{table}[t]
  \caption{\textbf{Real-robot task success rate} of \method and supervised learning approaches. Standard errors are calculated between groups of 10 consecutive roll-outs.}
  \label{tab:rl-sft}
  \centering
  \small
  \setlength{\tabcolsep}{3pt}
  \renewcommand{\arraystretch}{1.1}
  \begin{tabularx}{\columnwidth}{@{}Xccc@{}}
    \toprule
    Task & SFT & Iter-SFT & \method \\
    \midrule
    Open the drawer
      & $40 \pm 10\,\%$ & $30 \pm 10\,\%$ & $58 \pm 4\,\%$ \\
    Close the drawer
      & $62 \pm 6\,\%$ & $60 \pm 8\,\%$ & $62 \pm 6\,\%$ \\
    Put the eggplant into the blue sink
      & $4 \pm 4\,\%$ & $10 \pm 3\,\%$ & $72 \pm 10\,\%$ \\
    Put the eggplant into the yellow basket
      & $8 \pm 4\,\%$ & $17 \pm 17\,\%$ & $78 \pm 2\,\%$ \\
    \bottomrule
  \end{tabularx}
\end{table}

\paragraph{Comparing \method to Supervised Learning.}
We also compare \method with supervised fine-tuning methods. Following the recipe of OpenVLA-OFT ~\cite{kim2025fine}, we fine-tune a OpenVLA 7B policy ~\cite{kim2024openvla} on expert trajectories from the Bridge V2 dataset ~\cite{walke2023bridgedata} for 20k steps. This policy, denoted as \textit{SFT} in Table~\ref{tab:rl-sft}, is also the base model on which we conduct RL training. Recent works like Ctrl-World ~\cite{guo2025ctrl} further utilize the world model for policy improvement by generating synthesized roll-outs filtered by a reward model as additional supervision. Similar to Ctrl-World, we roll out the base \textit{SFT} policy in our world model for the same amount of steps as \method did for RL training, and filtered for successful trajectories on OpenVLA evaluation tasks using a VLM. We then conduct another iteration of supervised fine-tuning with a mixture of data from Bridge V2 plus the successful synthesized roll-outs with a $1:1$ sampling rate. The resulting policy, denoted \textit{Iter-SFT} in Table~\ref{tab:rl-sft}, is then evaluated on real world held-out tasks through AutoEval.

As shown in Table~\ref{tab:rl-sft}, \method achieves the best performance compared with supervised learning and Iter-SFT, with a 18-fold and nearly 10-fold improvements from the base policy on \emph{Put the eggplant into the blue sink} and \emph{Put the eggplant into the yellow basket}, respectively. Notably, Iter-SFT improves slightly on the harder tasks, but the performance degrades on the easier ones. One possible explanation is that RL, through active exploration and on-policy updates, learns more generalizable behaviors. In contrast, iterative SFT may overfit to synthetic experience and is more vulnerable to world model hallucinations and inaccurate VLM success judgments.

\subsection{Evaluating Diverse Settings \method Offers}\label{sec:exp-diverse}

\paragraph{Evaluating Training with Distractors.} We use Nano Banana \citep{SharonEtAl2025GeminiImageEditing} to generate a new dataset using the pre-existing frames from the OpenVLA Bridge task suite. The new dataset adds random objects to distract the policy from successfully achieving the given task. We then train a new policy with RL (\method-Distract) by including these new frames in the training data. Next, we evaluate the performance of SFT, \method and \method-Distract on a held-out set of distractor frames. WorldGym evaluations show \method-Distract is the most robust, while SFT is the easiest to distract (Figure~\ref{fig:distractor-qualitative}). Additionally, we evaluate \method-Distract on the original OpenVLA tasks in WorldGym and observe improved success rates (Table~\ref{tab:worldgym}). This indicates that adding distractor-augmented data improves performance not only under visual perturbations but also on the original tasks. Qualitative rollout comparisons in Figure~\ref{fig:distractor-qualitative} further illustrate that under visual distractions, World-Gymnast-Distract executes more reliable grasping and placement behaviors than SFT and World-Gymnast, while maintaining correct object grounding despite visual artifacts from the world model.

\paragraph{Evaluating Training with Novel Language Instructions.} Another approach to scaling data is augmenting language instructions in pre-existing tasks. We test this approach by creating 4 new tasks involving new interactions with objects already present in the scene. We combine this new data with the OpenVLA dataset and train a new policy, \method-Language. Next we evaluate the performance of \method-Language on the held-out split from OpenVLA data and observe that \method-Language has improved performance over \method (Table~\ref{tab:worldgym}). This suggests that creating more tasks by introducing novel language instructions on existing frames can further improve the generalization performance of VLA policies.

\begin{table}[t]
  \caption{\textbf{Comparing RL on diverse settings.} Leveraging the diverse capabilities of world modeling, \method allows significant improvement in the task success rates for multiple settings.}
  \label{tab:worldgym}
  \centering
    \begin{tabular}{lc}
\toprule
        Variant & Success Rate \\
        \midrule
        SFT & $58 \pm 4\,\%$ \\
        \method & $74 \pm 3\,\%$ \\
        \method-Distract & $78 \pm 2\,\%$ \\
        \method-Language & $81 \pm 1\,\%$ \\
        \method-Scaled & $81 \pm 4\,\%$ \\ 
        \bottomrule
\end{tabular}
\end{table}

\paragraph{Scaling the Number of Training Tasks.} One advantage of \method is its ability to train on a diverse set of tasks starting from any initial frame. To scale up data, we randomly selected 5 additional tasks from the Bridge dataset \citep{walke2023bridgedata}. We then train on these tasks in addition to the OpenVLA tasks and call this variant \method-Scaled. We evaluate the performance of \method-Scaled in WorldGym on the OpenVLA held-out task split and observed improvement compared to \method, as shown in Table~\ref{tab:worldgym}. These results suggest that \method can effectively leverage additional training tasks to improve performance.

\subsection{Evaluating Test-Time Optimization}\label{sec:exp-inference}

Pretrained policies often struggle at generalizing well to novel real world scenarios. While online data collection followed by finetuning can address this gap, it is prohibitively expensive in terms of time and effort. With a pretrained world model, we show that \method improves the performance of a base policy through test-time training without real world roll-outs. Specifically, provided only with the initial observation and the task instructions of the 4 scenarios from AutoEval \citep{zhou2025autoeval}, we fine-tune our base policy with RL (details in Appendix~\ref{app:hyperparams}) using imagined roll-outs generated by the world model in a zero-shot manner from the testing frame. Test time training significantly improves the performance and robustness of \textit{close the drawer} in the real world, improving the success rate from $62 \pm 6\%$ success rate to $100 \pm 0\%$ for the \emph{Close the drawer} task. However, we noted that test-time training overfits the model to this single task, and performance in other tasks generally degrade. Test-time training across diverse tasks is an interesting venue of future work.

\subsection{Evaluating Iterative World and Policy Improvement} \label{sec:exp-dyna}

\begin{figure}[t]
    \centering
    \includegraphics[width=\columnwidth]{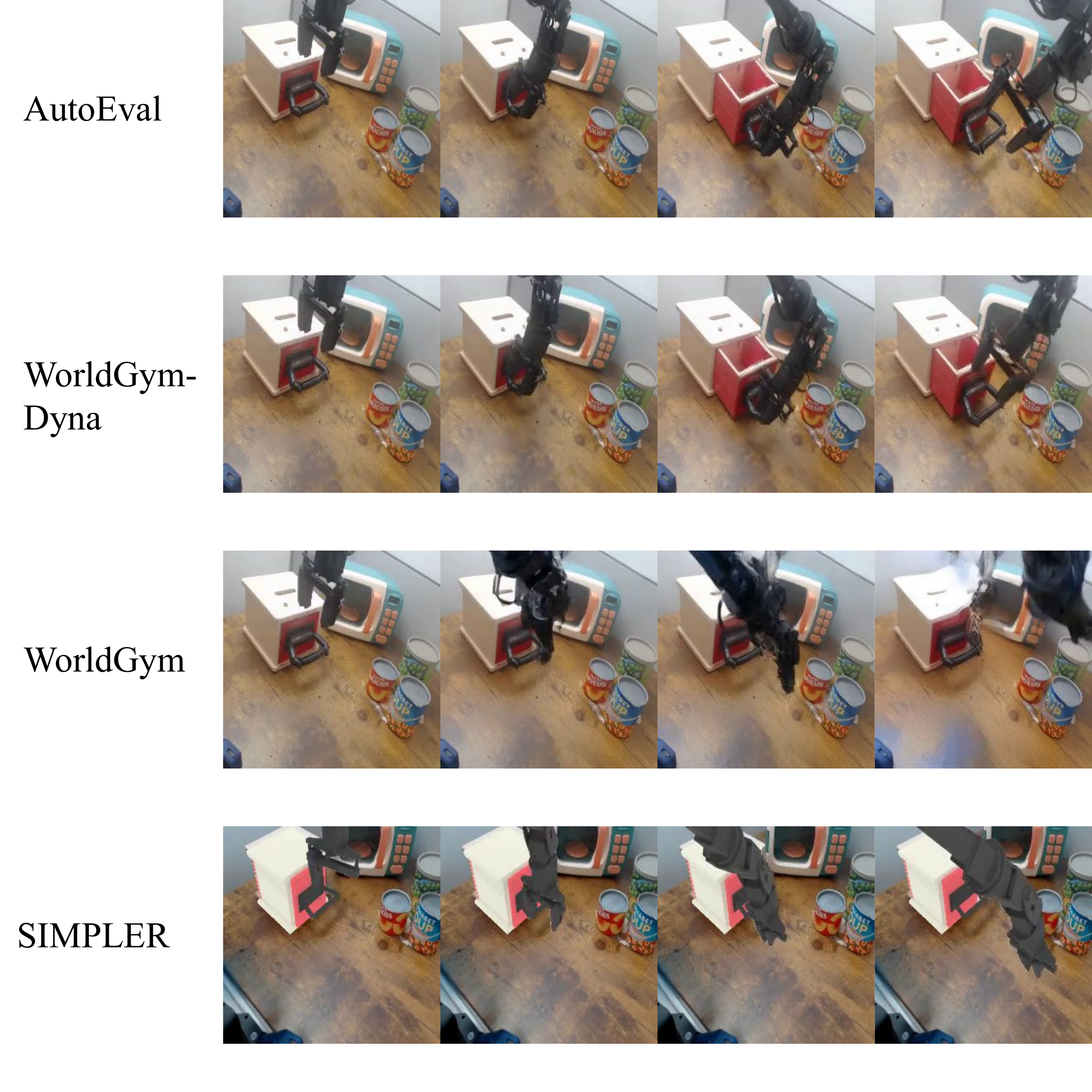}
    \caption{\textbf{Qualitative comparison} of rolling out the same action sequence on the real robot from AutoEval~\citep{zhou2025autoeval}, from software simulator SIMPLER~\citep{li2024evaluating}, from WorldGym~\citep{quevedo2025worldgymworldmodelenvironment}, and from \method with online world model updates. Rollouts from \method adheres more closely to the real world than SIMPLER, suggesting improving the world model through Dyna~\citep{sutton1991dyna} improves the quality of the rollout.}
    \label{fig:simpler-qualitative}
\end{figure}

A unique advantage of \method is that the world model can be iteratively updated with real world roll-outs for initial frames that are out-of-distribution of the pretrained world model. During policy evaluation with AutoEval \citep{zhou2025autoeval}, we save the task instructions, observations, and action sequences to improve our world model. Iteratively, we collected around 100 trajectories per task for all 4 tasks, on which we finetune the world model for a total of 120k steps. Figure~\ref{fig:simpler-qualitative} provides a qualitative result demonstrating the improvement of the world model in playing a sequence of actions for \emph{Open the drawer} task, which shows much less sim-to-real gap than SIMPLER~\citep{li2024evaluating} and than WorldGym without online updates. Using the Dyna style world model updated with online data as an RL environment, \method improves the success rate of the base RL model for \textit{close the drawer} in AutoEval to $95\%$.

%% file: related.tex
\section{Related Work}

\paragraph{Model-Based Reinforcement Learning.} Model-based RL has long been studied in the RL literature, which learns a dynamics model from previously collected data and rolling out the learned dynamics model for policy evaluation and improvement~\citep{sutton1991dyna,tani1996model,ljung1994modeling,liu2019reinforcement,zhang2021autoregressive,yu2020mopo,hafner2020mastering}. Much of the model-based RL research has been focusing on learning one dynamics model per system in the lower dimensional state space as opposed to in the pixel space~\citep{ferns2004metrics,achille2018separation,lesort2018state,castro2020scalable}, which, despite being a simpler modeling problem, limits knowledge sharing across systems. With large transformer architectures, learning image-based world models followed by RL has become plausible~\citep{hafner2020mastering,chen2022transdreamer,seo2022reinforcement,micheli2022transformers,wu2022slotformer,hafner2023mastering}, but mostly in games or simulated domains with visually simplistic and abundant data. Our work differs from existing model-based RL in that we focus on using a single world model learned on broad data from many policies and tasks to train a generalist VLA policy on novel language instructions and scenes. We also directly tackle the sim-to-real gap by comparing against RL in traditional simulators, further showing the value of model-based RL where the model is learned from real-world data.

\paragraph{Sim-to-Real RL.}
Reinforcement learning in physics-based simulations has been widely adopted to overcome the sample inefficiency of real-world training. To bridge the reality gap, prior works have relied heavily on domain randomization ~\citep{tobin2017domain,peng2018sim}, which varies visual and physical parameters to cover real-world distributions, or domain adaptation \citep{bousmalis2018using,rao2020rl}. These strategies have achieved notable success in locomotion~\citep{tan2018simtoreallearningagilelocomotion,rudin2022learningwalkminutesusing} and rigid-body manipulation~\citep{openai2019solvingrubikscuberobot,handa2024dextremetransferagileinhand}. However, traditional simulators~\citep{conf/iros/2012,makoviychuk2021isaacgymhighperformance} face a scalability bottleneck for generalist manipulation: they require explicit object modeling, manual scene engineering, and struggle to faithfully render the diverse visual textures and deformable dynamics of the real world. Unlike these approaches, \method leverages a world model learned directly from real-world data, effectively bypassing the need for manual asset creation and physics parameter tuning.

\textbf{Video Generation for Robot Learning.} Video-based learning for robotics ~\citep{nair2022r3m,bahl2022human,shao2021concept2robot,chen2021learning,pari2021surprising} has enabled visual representation learning, goal extraction, planning ~\citep{finn2017deep,kurutach2018learning}, and imitation from expert actions ~\citep{fang2019survey, wang2023diffusion, DiffClone}. Recent works reframe decision-making as a text conditioned video generation task, enabling policy learning from video predictions~\citep{du2024learning, ko2023learning, wen2023any, du2023video, ajay2024compositional}, and use generative models to simulate agent-environment interactions ~\citep{yang2023learning}. Most of these work use generated video plans as visual actions and train separate inverse dynamics to extract robot actions from generated videos. While text-to-video generation can be effective for long-horizon planning, it is less clear how to self-improve these models that use video-generation as policies. We study the problem of using video generation solely as environment and using RL with generated rollouts to improve policy performance. Notably, \method in principle can be used to improve any policies beyond VLA policies, including policies that are parametrized through video generation.

\paragraph{Policy Evaluation using World Models.} Recent work have shown that world models learned from real-robot data can approximate real-robot execution outcomes, and hence be used to evaluate robot policies~\citep{quevedo2025worldgymworldmodelenvironment,guo2025ctrl,tseng2025scalable,li2025worldeval}. While evaluating policies in a world model is by all means an important application of world models, we focus on the end-to-end problem of improving policy performance using a world model, the effect of which can be tested on real robot.

\paragraph{RL with a Video Based World Model.} Our work is the most similar to \citet{zhu2025wmpo} which uses RL to improve a policy in a video world model, but we focus on real-world evaluation with easily accessible evaluation settings using a open-source VLA policy (OpenVLA), world model (WorldGym), and evaluation platform (AutoEval). We also focus on exploring the emergent capabilities of RL in a world mdoel, including training from any initial frame with novel language instructions, test-time training, and iterative world model and policy improvement.

%% file: conclusion.tex
\section{Conclusion and Limitations}
We have presented \method, an RL framework for fine-tuning VLA policies using a learned world model. We show that existing training paradigms, including SFT and software-simulator-based RL, are expensive, restrictive, and often produce policies with limited generalization. In contrast, using a world model to rollout policies and VLM for reward is cheap, scalable and more robust to OOD scenarios.

A key advantage of world-model-based training is the ability to generate diverse training data from minimal inputs. Since \method requires only an initial scene and a task description, it naturally supports data augmentation through image editing, language variation, and the reuse of scenes from novel environments. This flexibility further enables test-time training, planning, and iterative improvement of both the policy and the world model.

\paragraph{Limitations.} One limitation of \method is that it cannot generalize to an arbitrary initial frame if the initial frame is far from the world model's training distribution. This calls for future research of pretraining robot world models on broad robot datasets. Another limitation of \method is the reliance on a pretrained VLM for task success, which can produce hallucinations that leads to suboptimal RL training. Exploring ways to improve the reward model, such as training reward models~\citep{lee2026roborewardgeneralpurposevisionlanguagereward}, is an important future direction. Furthermore, utilizing dense rewards from VLMs and preventing reward hacking are also promising directions for future work.